# DualVAE : Controlling Colours of Generated and Real Images

KEERTH RATHAKUMAR, University of New South Wales, CSIRO's Data61, and Cyber Security CRC, Australia
DAVID LIEBOWITZ, Penten, Australia
CHRISTIAN WALDER, Google Brain, Canada
KRISTEN MOORE, CSIRO's Data61 and Cyber Security CRC, Australia
SALIL S. KANHERE, University of New South Wales, Australia

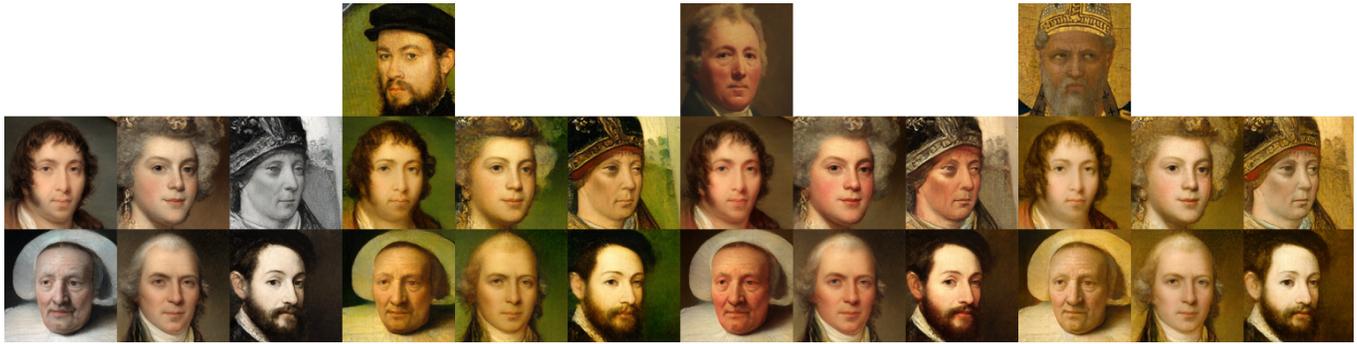

Fig. 1. Example DualVAE image generation results on Metfaces: The leftmost 2 × 3 block shows images generated in the unconditional setting (i.e. both geometry and colour latents are sampled from their prior). The remaining three 2 × 3 blocks correspond to conditional generation, where the colours of each 2 × 3 block are derived from its exemplar image directly above. In all three blocks, the same six geometric tokens are used.

Colour controlled image generation and manipulation are of interest to artists and graphic designers. Vector Quantised Variational AutoEncoders (VQ-VAEs) with autoregressive (AR) prior are able to produce high quality images, but lack an explicit representation mechanism to control colour attributes. We introduce DualVAE , a hybrid representation model that provides such control by learning disentangled representations for colour and geometry. The geometry is represented by an image intensity mapping that identifies structural features. The disentangled representation is obtained by two novel mechanisms: *(i)* a dual branch architecture that separates image colour attributes from geometric attributes, and *(ii)* a new ELBO that trains the combined colour and geometry representations. DualVAE can control the colour of generated images, and recolour existing images by transferring the colour latent representation obtained from an exemplar image. We demonstrate that DualVAE generates images with FID nearly two times better than VQ-GAN on a diverse collection of datasets, including animated faces, logos and artistic landscapes.

## 1 INTRODUCTION

Deep generative models have achieved significant success in image synthesis tasks, such as generating realistic images, image-to-image translation, colour and style transfer and text-guided synthesis. These models are of particular interest in the artistic domain due to their ability to synthesise impressive artworks. While the generation of high-fidelity artwork has been well studied, generating artworks with desired attributes remains an open problem and an interest to the artistic domain. In particular, artists are interested in specifying the colours to an artwork generator. This may be as simple as specifying the colours of the generated artwork to be derived from a colour palette.

Much of the work on colour control has focused on on editing colours of existing images through pixel shifts [Afifi et al. 2019; Avi-Aharon et al. 2020; Chang et al. 2015; Laffont et al. 2014; Nguyen et al. 2014] or using deep learning models [Deshpande et al. 2017; Huang et al. 2022; Kumar et al. 2021; Luan et al. 2017; Saharia et al. 2022; Su et al. 2020; Vitoria et al. 2020], but the study of colour controllable generative models are limited to StyleGAN and HistoGAN [Afifi et al. 2021; Karras et al. 2020b].

StyleGAN[Karras et al. 2020b] and HistoGAN[Afifi et al. 2021] are Generative Adversarial Networks (GANs) [Goodfellow et al. 2014] that can perform colour controllable generation. However, GANs trained with an adversarial loss can be unstable to train and succumb to mode collapse [Arjovsky et al. 2017]. In contrast, Variational AutoEncoders (VAEs) and VQ-VAE [Chang et al. 2022; Esser et al. 2021b,a; Ramesh et al. 2021; Razavi et al. 2019; Van Den Oord et al. 2017] are likelihood based generative models that may be more reliably used due to their more stable model training and greater mode coverage [Sajjadi et al. 2018; Zhong et al. 2019].

VAEs have a latent variable that drives the attributes of the generated content [Kingma and Welling 2013; Rolinek et al. 2019]. This latent representation enables VAEs to mix attributes from one image to another to perform attribute transfer [Zhang et al. 2019]. Conditional VAEs (cVAE) with a U-Net [Ronneberger et al. 2015] architecture are of particular importance to appearance transfer. They disentangle appearance when conditioned on a geometry estimate [Deshpande et al. 2017; Esser et al. 2018; Wu et al. 2019]. In particular, they can perform colour transfer and recolouring

Authors' addresses: Keerth Rathakumar, University of New South Wales, CSIRO's Data61, and Cyber Security CRC, Australia, k.rathakumar@unsw.edu.au; David Liebowitz, Penten, Australia, david.liebowitz@penten.com; Christian Walder, Google Brain, Canada, cwalder@google.com; Kristen Moore, CSIRO's Data61 and Cyber Security CRC, Australia, kristen.moore@data61.csiro.au; Salil S. Kanhere, University of New South Wales, Australia, salil.kanhere@unsw.edu.au.

when a grayscale transformed image is given as the geometry estimate [Deshpande et al. 2017; Esser et al. 2018]. In both cases, the geometry of the image as identified by the grayscale transform is unchanged, indicating that the learnt colour latent variable is disentangled from geometry.

Instrumental to obtaining the disentangled latent variable of U-Net cVAEs is their skip connections, which helps preserves the spatial information by propagating it directly to the output [Esser et al. 2018]. These skip connections ensure that the information from the structure estimate is used prior to the latent variable [Deshpande et al. 2017; Esser et al. 2018; Wu et al. 2019]. This tends to bias the latent variable of the cVAE model to learn the residual information not present in the structure estimate as its latent variable. For example, using edges or grayscale as structure leads to learning colours[Deshpande et al. 2017; Esser et al. 2018], while using as face landmarks lead to learning identity [Qian et al. 2019; Wu et al. 2019], and appearance when human pose estimate is given [Esser et al. 2018]. We focus our attention on those that use grayscale or edge images as the structure estimate, resulting in a model that learns colour latent variables.

While the cVAE model has a latent variable for colours from which we could sample new colours, they do not have a latent variable for its geometry representations. This limits their usability to the artistic domain because they could only edit colours of an existing artwork rather than generate entirely new artwork with a desired colour. However, if we were to map the structure estimate to another latent variable but retain a suitable skip connection architecture that disentangles, we would obtain disentangled latent variables for colour and structure. This would result in a colour controllable generative model of art because the geometry latent variable generates the overall structure of the art. On the other hand, suitable colour features for the model can be controlled by transferring colours from an exemplar artwork. However, learning two latent variables for geometry and colour is difficult due to their strong correlations in the training data, which leads one latent variable to overpower the other [Esser et al. 2019; Jha et al. 2018; Mathieu et al. 2016].

We introduce DualVAE , a variational autoencoder that uses skip connections and a novel regularization that allows it to learn disentangled latent variables for geometry and colour features. DualVAE achieves disentanglement of geometry from colours by building on the variational U-Net architecture, which has been shown to obtain a disentangled colour latent variable due to the use of skip connection. However, DualVAE , unlike the existing variational U-Nets, contains a latent variable for image structures, allowing it to perform generation of images whose structure is not present in the training set. Our novel regularization enables learning of these two latent variables, which we show using an ablation avoids the degenerate case of one latent overwhelming the other. Additionally, we alleviate the need for a pretrained structure estimate by using our geometry module. The geometry module is an intensity transform that begins with a sequence of 3x3 convolutions that keeps the spatial resolution the same, ending with a layer that performs channel-wise averaging . This module suppresses colours by aggregating intensity channel-wise, much like grayscale. However, grayscale transform can lose pixel boundaries by greedily aggregating RGB to one channel. This is avoided by our geometry module, which learns an adaptive intensity transform suited to the data and model by incorporating it as part of back-propagation. An example output from the geometry module is shown in Figure 2. The colour latent in a cVAE is typically a Gaussian [Esser et al. 2018]. However, the structure estimate has spatial information, which is localized and cannot be encoded in a continuous latent. Instead, we use a vector quantized token latent that can learn localized spatial features by quantising them to a spatial token grid. An autoregressive prior is then learnt over the token grid.

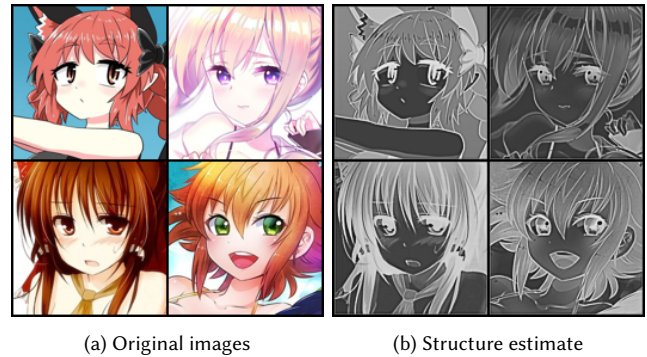

(a) Original images         (b) Structure estimate

Fig. 2. Example of structure estimate learnt by DualVAE . Learnt structure representation exaggerates the intensity at the pixel boundaries, behaving similar to edge detectors. However, unlike classical edge detectors, our structure estimate does not introduce noise.

### 1.1 Contribution

Our main contributions are as follows:
- Architectural changes to cVAE that biases VAE to learn disentangled latent variable for colours and geometry.
- Novel regularization that stabilizes the learning of the latent representations of colour and geometry. This enables DualVAE to perform various generative tasks;
  - Unconditional generation: Both colour and geometry latents are sampled and decoded to generate new images.
  - Colour controlled generation by providing an exemplar: The geometry latent is sampled, but the colour latent is obtained by passing the exemplar through the colour encoder.

We demonstrate on several datasets that the combination of architectural changes and our novel regularization strongly biases the model toward a meaningful separation of colour and geometry[1]. In addition, we show that the generative performance of DualVAE exceeds or is comparable to its baseline model VQ-GAN. We also introduce a variant of DualVAE, which we call ReDualVAE, that is more suitable for manipulating colours of existing images. For the interested reader the ReDualVAE is detailed in the supplementary, but the main focus of the paper is colour controlled generating of images rather than recolouring existing images.

---
[1]Code will be released upon publication

## 2 RELATED WORKS

### 2.1 The Varional AutoEncoder (VAE)

A Variational AutoEncoder is a likelihood model, which we parameterise with $\theta$, of observations $X$ and an unobservable latent variable $z$ whose prior distribution is $p(z)$. It is learned by maximising the model evidence $p_\theta(X)$. However, obtaining the model evidence $p_\theta(X)$ requires marginalising the unobserved latent variable $z$ from the joint density of $(X, z)$, which is typically an intractable task. We instead optimise the Evidence Lower Bound (ELBO), a lower bound of the model evidence.

The ELBO of a VAE is shown in Equation (1) and is maximised using an encoder-decoder consisting of neural networks:

$$p_\theta(X) \geq \mathbb{E}_{q_\phi(z|X)}\left[p_\theta(X|z)\right] - \mathbb{KL}\left[q_\phi(z|X) \mid p(z)\right] \quad (1)$$

The observations $X$ are passed to the encoder with parameters $\phi$ to obtain their approximate posterior distribution $q_\phi(z|X)$ and a decoder is used to maximise $p_\theta(X|z)$.

The approximate posterior distribution $q_\phi(z|X)$ is typically assumed to be a conditional Gaussian distribution with diagonal covariance as given by $q(z|X) = \mathcal{N}\left(\mu_\phi(X), \text{diag}\left(\sigma_\phi(X)\right)\right)$. The decoding distribution $p_\theta(X|z)$ is typically Gaussian with identity covariance $p_\theta(X|z) = \mathcal{N}(\mu_z, I)$.

### 2.2 Conditional Generative Models For Editing Appearance

The literature on editing appearance includes VAEs [Deshpande et al. 2017; Esser et al. 2018; Qian et al. 2019; Wu et al. 2019], normalising flows [Li et al. 2022] and GANs [Cao et al. 2017b]. They are conditional generative models that use structure (typically edges) as the conditioning information to transfer appearance from an exemplar images to the structure image.

The structure estimates are typically outputs from existing pretrained deep learning models and include sketches [Xie and Tu 2015], landmarks [Zhang et al. 2018a] and pose [Cao et al. 2017a]. These models are not able to perform generation, but instead transfer appearance from an exemplar image to a source shape estimate [Esser et al. 2018; Wu et al. 2019]. In contrast, DualVAE is a generative model of both structure and colours. This means DualVAE is able generate structure not present in the training data, unlike these conditional methods.

### 2.3 Vector Quantised Variational AutoEncoder (VQ-VAE)

Autoregressive (AR) models of images are likelihood models in pixel space. However, they are challenging to train directly because of the large dimensionality of images.

VQ-VAE circumvents this by first reducing images to a smaller spatial grid in which training autoregressive models is less challenging. An AR generative model is then trained in a second stage. Generation is performed by first sampling a token sequence and passing it to the VQ-VAE decoder. Training such AR models on the token sequence and sampling is computationally expensive, so transformer AR models are more suitable at larger resolutions. However, these models tend to generate images with inconsistent global structure [Esser et al. 2021a].

### 2.4 VQ-GAN

VQ-GAN uses a patch-based discriminator [Isola et al. 2017] to learn a more suitable token sequence for transformer based models. This allows transformer models to generate images with more consistent global structures [Esser et al. 2021a] . It can also perform conditional modelling of images by prepending conditioning information to the token sequence used by the transformer AR model. The conditioning information is used during both training and generation. However, if the conditioning information has spatial characteristics, then an additional VQ-GAN is learnt to obtain the token sequence of the spatial characteristics. Then transformer model is trained on the token sequence obtained by prepending this additional token sequence to the original token sequence.

### 2.5 HistoGAN

HistoGAN, a generative adversarial model, was recently introduced as a colour controllable generative model [Afifi et al. 2021]. The idea is to construct an architecture that is colour controllable by conditioning StyleGAN [Afifi et al. 2021; Karras et al. 2020b] with colour histogram features. Transferring colours from target to source image is achieved by transferring the histogram feature of the target to the source in the generative step [Afifi et al. 2021]. However, GANs are known for their training instability and do not cover all modes of the data [Arjovsky et al. 2017]. DualVAE is a variational inference model, and thus has a (stochatic) inverter [Nielsen et al. 2020] , better model training stability than the GAN, and does not succumb to mode collapse. Additionally, unlike HistoGAN, DualVAE is able to colourise grayscale images without requiring the user to provide colour inputs.

## 3 METHODOLOGY

In this section we provide an overview of the DualVAE architecture, decoding and regularization steps. We then interpret and justify the regularization step as an implicit ELBO, which we derive using variational inference. Finally, we present a simpler variant of the DualVAE model architecture, ReDualVAE, for manipulating colours in existing images.

### 3.1 Overview of the Design

DualVAE performs a sequence of changes to the U-Net cVAE architecture shown in Figure 4a to obtain a colour controllable generative model. We introduce two latent representations, a geometry latent $z_g$ and colour latent $z_c$. The architectural changes are made to the decoder section and are illustrated in Figure 4b. These include adding separate skip connections from $z_g$ and $z_c$ and a merge pathway.

The architectural changes are shown in Figure 4b and are summarised below:

- **Introduce latent $z_g$**: We first modify the cVAE architecture to introduce a latent for the geometry estimate. While a continuous latent space is used for colours, it is not appropriate for geometry features. Geometric features are localized and easily get lost in the continuous bottleneck [Esser et al. 2018]. However, a token embedding is able to map the geometry estimate to a spatial grid of tokens, each learning features in its local region.

- **Skip connections from $z_g$**: Encoding the geometry estimate to the latent $z_g$ results in the removal of the disentangling skip connections shown in Figure 4a. We reintroduce those skip connections from $z_g$ to DualVAE to the decoder (shown in Figure 4b). However, these skip connections require some form of feature normalisation for stable training [Vaswani et al. 2017; Wu and He 2018], so we introduce layer normalisation [Ba et al. 2016] after each skip connection.
- **Skip connections from $z_c$**: While the layer normalisation helps stabilize training, it blocks signal propagation from the latent, so we also introduce skip connections from $z_c$.
- **Decoder**: To attain signal propagation similar to cVAE, signals from $z_g$ needs to be used earlier in the reconstruction of $X$. We achieve this signal propagation by using the merge node shown in Figure 3.

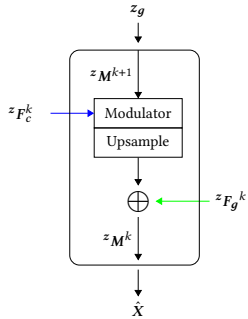

Fig. 3. Detailed view of the decoding pathway of Figure 4b. Each block of this figure corresponds to a merge module with $k$ representing its ordering in the decoding pathway and the ${}^zM^k$ denotes its input. Also, we denote outputs of the skip connections from $z_g$ and $z_c$ shown in Figure 4b as ${}^zF_g{}^k$ and ${}^zF_c{}^k$ respectively. Merge module combines colour features ${}^zF_c{}^k$, performs upsampling and concatenates geometry features ${}^zF_g{}^k$. This propagates signals from geometry latent $z_g$ before $z_c$.

These changes result in an additional loss term for $z_g$ to the cVAE loss:

Recons loss = $\left\| X - D_X({}^zF_g, {}^zF_c) \right\|_1$

Latent loss = (VQ Latent Loss of $z_g$) + (KL Latent Loss of $z_c$)

where $D_X$ denote the merge pathway consisting of modules $\boxed{M}$ shown in Figure 4b. Also, ${}^zF_g, {}^zF_c$ is used to summarise the combined skip connection outputs $[{}^zF_g^k]$ and $[{}^zF_c^k]$.

### 3.2 Regularization: Training two latent variables

*3.2.1 Latent variable degeneracy.* The model described above cannot be trained by minimizing the model log-likelihood alone. In particular, there is nothing to prevent all of the information about the images from flowing through one latent variable. This may lead the decoder to use only one latent variable to decode images, leaving the other unused. This degenerate solution can be prevented by regularizing the decoder [Georgopoulos et al. 2020; Mathieu et al. 2016]. We introduce a novel regularization that avoids this degenerate case.

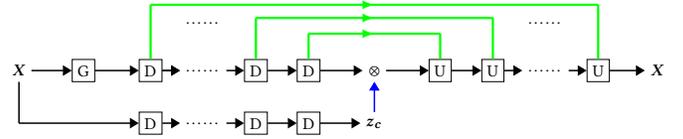

(a) Conditional VAEs (cVAE). Skip connections from structure estimates ensure that the latent $z_c$ is disentangled.

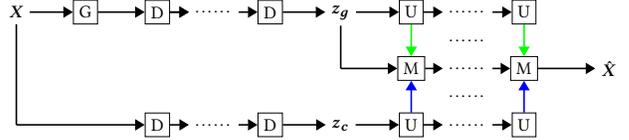

(b) Introducing the additional latent $z_g$ to the cVAE architecture leads to the loss of colour disentangling skip connections shown in Figure 4a. Thus DualVAE, introduces a decoding pathway consisting of merge modules $\boxed{M}$ and skip connections. The merge modules $\boxed{M}$ use the skip connections from $z_g$ prior to using those from $z_c$, which results in similar signal propagation to cVAE and a disentangled $z_c$.

Fig. 4. Schematic overview of cVAE and DualVAE: $\boxed{U}$ and $\boxed{D}$ are (U)psampling and (D)ownsampling convolutions. Features used by the decoder section to reconstruct input $X$ are shown in blue or green. Blue lines indicate the positions where the features that depend only on $z_c$ are injected, and green indicates injection of features dependent only on the structure estimate. Both DualVAE and U-Net cVAE ensure that at each feature hierarchy, the feature outputs that describe structural features are available earlier than those that depend on $z_c$.

*3.2.2 Regularization.* Our regularization is an additional reconstruction task for the decoder, where the outputs of skip connection from a latent are replaced with feature outputs obtained by the encoder of that latent. Specifically, we replace ${}^zF_g{}^k$ and ${}^zF_c{}^k$ shown in Figure 3 with intermediate features obtained by colour and geometry encoders. $F_g{}^k$ and $F_c{}^k$ are the intermediate features obtained by the colour and geometry encoders. We note that the superscript $z$ is used to differentiate between outputs of skip connections (E.g. ${}^zF_g$) and the intermediate features of the encoders (E.g. $F_g$).

This leads us to the loss of DualVAE, the sum of the component losses;

Reconsloss = $\left\| X - D_X({}^zF_g, {}^zF_c) \right\|_1$ (2)

Latentloss = (VQ Latent Loss of $z_g$) + (KL Latent Loss of $z_c$)

Regularizationloss = $\left\| X - D_X(F_g, F_c) \right\|_1$.

We demonstrate that without this regularization loss only the geometry latent is used by the decoder to generate images, resulting in a lack of colour control.

*3.2.3 Ablation.* We compare color control with and without regularization by performing an ablation study that examines the deviation between colour features of generated images and exemplar images. We measure the colour deviation by extracting histogram features [Afifi and Brown 2019] of generated and exemplar images, and compute Kullback–Leibler divergence between them [Afifi et al.

2021]. We performed the evaluation on test portion (5000 images) of Animated Faces dataset [Anonymous et al. 2021].

The results of our ablation study is shown in Table 1. A lower value of $\mathbb{KL}$ divergence indicates that the colours of generated images are close to the colours of exemplar images.

| Model Run | $\mathbb{KL}$ Divergence ↓ |
|---|---|
| w/o Regularization | 0.9408 |
| with Regularization | **0.6834** |

Table 1. Ablation study of exemplar guided colour control on $64 \times 64$ images: The degree of colour control is measured using $\mathbb{KL}$ divergence between histogram features of generated and exemplar images. Average histogram $\mathbb{KL}$ divergence between any two test images is 0.9799, close to the model run without regularization. Lower is better.

### 3.3 Motivating the DualVAE Architecture from a Variational Inference Perspective

In this section we motivate the design of DualVAE [2] by interpreting the skip connections from the latent as decoders in their own right, tasked with decoding features obtained during encoding. In light of interpreting the skip connections as decoders, we re-draw the diagram in Figure 4b, but replace the skip connections from $z_g$ and $z_c$ as decoders $D_G$ and $D_C$. Also, for the sake of brevity, we denote Encoder $G$ and Encoder $C$ as $E_G$ and $E_C$. This results in Figure 5.

The regularization loss described earlier can then be interpreted as a reconstruction loss of $D_G$ and $D_C$. Training $D_X$ with both the regularization loss $\left\|X - D_X(F_g, F_c)\right\|_1$ and reconstruction loss $\left\|X - D_X(^zF_g, ^zF_c)\right\|_1$ forces $F_g, F_c$ to be close $^zF_g, ^zF_c$. This is due to $D_X$ taking them as inputs to reconstruct $X$, which forces those inputs to be close to each other. This encourages both latent spaces to learn useful information.

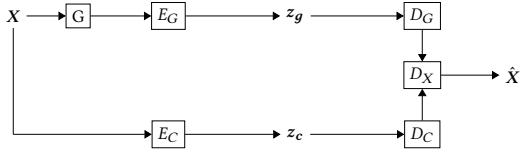

Fig. 5. DualVAE design in Figure 4b re-drawn but with skip connections from $z_g$ and $z_c$ re-written as $\boxed{D_G}$ and $\boxed{D_C}$. Also, we re-draw the the sequence of merge modules $\boxed{M}$ as $\boxed{D_X}$. DualVAE can be interpreted as consisting of two encoders and decoders, whose outputs are combined by the $D_X$ to generate images.

The loss in Equation 2 expressed when expressed more formally is

$$\mathcal{L}_{\text{DualVAE}} \stackrel{\text{def}}{=} -2 \cdot \mathbb{E}_{q_{\phi_c}(z_c|F_c) \cdot q_{\phi_g}(z_g|F_g)} \left[\left\|X - D_X(F_g, F_c)\right\|_1\right]$$
$$- \mathbb{E}_{q_{\phi_c}(z_c|F_c) \cdot q_{\phi_g}(z_g|F_g)} \left[\left\|X - D_X\left(D_G(z_g), D_C(z_c)\right)\right\|_1\right]$$
$$- \mathbb{KL}[q_{\phi_g}(z_g|F_g) || p(z_g)] - \mathbb{KL}[q_{\phi_c}(z_c|F_c) || p(z_c)]. \quad (3)$$

---
[2] A full mathematical derivation of DualVAE loss is given in the supplementary document

### 3.4 Image Generation

The DualVAE sampling process has similarities to VQ-VAE, where a token sequence $z_g$ is sampled using an autoregressive prior such as a transformer. However, DualVAE also has an additional colour latent representation from which we must sample.

*3.4.1 Unconditional.* The unconditional generation stage first samples the token latent $z_g$ from an autoregressive prior and a continuous latent $z_c$ from $\mathcal{N}(0, I)$. The latent representations are then passed to the decoder to generate new images.

The colours of the generated images can still be controlled in the unconditional setting. This is done by fixing the colour latent $z_c$ and varying the token latent $z_g$. However, the colours of the generated images are not predictable. To obtain specific colours, we need an exemplar which contains the colours we want to generate.

*3.4.2 Conditional.* Colour conditioned generation differs from the unconditional setting in that the colour latent representation used at generation stage is derived from an exemplar image instead of sampling from a prior. This ensures that the colours of the generated images are similar to the exemplar. We derive the colour latent representation $z_c$ by passing the exemplars through the Encoder C shown in Figure 4b. Finally, the $z_c$ and token latent $z_g$ are passed to the decoder to generate images with similar colours to the exemplar.

### 3.5 ReDualVAE : Manipulating Colours In Existing Images

ReDualVAE is a variant of DualVAE which has a latent space for colour features $F_g$, but not for geometry $F_g$. ReDualVAE is used to perform recolouring and colour transfer, but not generation. The ReDualVAE loss is derived by removing the token latent loss and reconstruction loss of $F_g$, leaving

$$\mathcal{L}_{\text{ReDualVAE ELBO}} \stackrel{\text{def}}{=}$$
$$- 2 \cdot \mathbb{E}_{q_{\phi_c}(z_c|F_c) \cdot q_{\phi_g}(z_g|F_g)} \left[\left\|X - D_X(F_g, D_C(z_c))\right\|_1\right]$$
$$- \mathbb{E}_{q_{\phi_c}(z_c|F_c) \cdot q_{\phi_g}(z_g|F_g)} \left[\left\|X - D_X\left(F_g, F_c\right)\right\|_1\right]$$
$$- \mathbb{KL}\left[q_{\phi_c}(z_c|F_c) || p(z_c)\right] \quad (4)$$

### 3.6 Reconstruction Loss

Our reconstruction loss function is the sum of $\mathcal{L}_1$ reconstruction loss and the Learnt Perceptual Image Patch Similarity (LPIPS) reconstruction loss. It is based on prior work showing improvements in generation attained by VAEs when LPIPS[3] loss is used [Hou et al. 2017; Zhang et al. 2018b]. In addition, DualVAE loss includes an adversarial perceptual loss [Esser et al. 2021a].

## 4 EXPERIMENTS

### 4.1 Datasets

We use Art Landscapes [Saleh and Elgammal 2015], Metfaces [Karras et al. 2020a], Anime Faces [Anonymous et al. 2021], Butterfly [But [n. d.]], Birds [Wah et al. 2011], and Logos [Sage et al. 2018] to evaluate the generative performance of DualVAE [4]. ReDualVAE is evaluated on ImageNet [Deng et al. 2009], Art [Saleh and Elgammal 2015], Flowers, Logos [Sage et al. 2018] and Landscapes [Skorokhodov et al. 2021]. The qualitative results of REDUALVAE are shown in Figure ?? and Figure 16 is shown in the figure only pages and quantitative results are shown in Table 4.

---
[3] https://github.com/richzhang/PerceptualSimilarity
[4] Train and test splits are shown in the appendix

## 4.2 Results

*4.2.1 Generation.* In Figure 6, we show examples of generated images obtained by colour conditioning DUALVAE with the exemplars in the top row. We observe in Figure 6 that DUALVAE is able to generate new images with colours derived from the exemplar. DUALVAE also demonstrates better generation compared to VQ-GAN[5] on several datasets as shown in Table 3.

Generative comparisons of vector quantized models typically use identical numbers of tokens and embedding dimension for each model. However, DUALVAE has an additional colour latent space, so model runs over identical codebook settings may not be a fair comparison. We demonstrate that increasing the number of tokens in VQ-GAN does not lead to improvements in image reconstruction.

We ran VQ-GAN with larger token sizes, evaluated their image reconstructions using Fréchet inception distance (FID) [Heusel et al. 2017] on 5000 images from test portion and report the results in Table 2. We observe that reconstructions of VQ-GAN does not improve for larger token sizes and obtains the best rFID at token size of 512 compared to DUALVAE with an rFID of 16.7. Therefore, in the model comparison shown in Table 3 we used VQ-GAN with 512 tokens.

| Model | 512 | 640 | 768 | 896 | 1024 | 2048 | 4096 |
|---|---|---|---|---|---|---|---|
| VQ-GAN | 33.4 | 39.4 | 43.6 | 40.6 | 39.5 | 39.6 | 43.6 |
| DUALVAE | **16.7** | | | | | | |

Table 2. VQ-GAN Reconstruction FID for different codebook size: Embedding dimension (dim=256) and down sampling factor ($f = 16$) are kept constant. Lower is better.

We compare the image generative performance using the most common evaluation metrics; precision, recall[Sajjadi et al. 2018], and Fréchet inception distance (FID) [Heusel et al. 2017]. We set up the model runs of DUALVAE and VQ-VAE with identical embedding dimension, number of tokens and a transformer prior with identical parameters. The precision, recall and FID of DUALVAE and VQ-GAN under this experimental settings are reported in Table 3. We observe that DUALVAE attains the best FID on all datasets. Furthermore, it attains the best precision and recall on all datasets except for butterfly and animated faces. It obtains a lower recall on animated faces and a lower precision on butterfly, but performs better overall.

*4.2.2 Editing Colours.* Our quantitative results is shown in Table 4. REDUALVAE is second to ReHistoGAN, except for Logos. This is due to the limitation of histogram feature as a colour descriptor.

---
[5]VQ-GAN Architecture https://github.com/CompVis/taming-transformers/

Logos typically have multiple shapes of the same size with different colours which cannot be described by a 2D histogram feature, resulting in a lower performance on logos dataset. However, performance of REDUALVAE exceeds ReHistoGAN and cVAE in grayscale colourisation task on the datasets evaluated.

## 5 CONCLUSION

We present DUALVAE, a novel Variational Autoencoder that differs from other models like cVAE and VQ-VAE in having identifiable and disentangled latent spaces for colours and geometry. This allows it to (1) perform colour controllable generation; (2) generate images with higher quality than VQ-GAN as measuserd by FID on several colour datasets; (3) Edit colours in existing images. We also introduce REDUALVAE, a variant of DUALVAE that demonstrates realistic colour manipulations on multiple data sets of varying levels of complexity. In particular, REDUALVAE is able to perform realistic recolouring of images and semantically meaningful interpolations through its colour latent space.

| Model | Dataset | Precision ↑ | Recall ↑ | gFID↓ |
|---|---|---|---|---|
| VQ-GAN | Birds | 0.869 | 0.928 | 31.83 |
| DUALVAE | | **0.945** | **0.959** | **13.76** |
| VQ-GAN | Logos | 0.907 | 0.953 | 49.55 |
| DUALVAE | | **0.989** | **0.973** | **25.46** |
| VQ-GAN | Anime Face | 0.897 | **0.967** | 51.23 |
| DUALVAE | | **0.951** | 0.937 | **27.97** |
| VQ-GAN | Butterfly | **0.981** | 0.979 | 23.93 |
| DUALVAE | | 0.976 | **0.981** | **12.01** |
| VQ-GAN | Art Landscapes | 0.849 | 0.912 | 65.46 |
| DUALVAE | | **0.951** | **0.975** | **28.73** |
| VQ-GAN | AFHQ | 0.951 | **0.967** | 30.41 |
| DUALVAE | | **0.986** | 0.964 | **18.04** |

Table 3. Generative comparisons on 128 × 128 images: For Embedding dimension (dim=256), codebook size ($N_{\text{Embed}} = 512$) and downsampling factor ($f = 16$). Precision, Recall and gFID (Generated FID) are used.

| Model | Dataset | Transfer FID↓ | Colourisation FID↓ |
|---|---|---|---|
| ReHistoGAN | ImageNet | **5.24** | 30.77 |
| cVAE | | 25.30 | 25.30 |
| REDUALVAE | | 13.24 | **15.20** |
| ReHistoGAN | Flowers | **8.95** | 67.29 |
| cVAE | | 14.60 | 13.99 |
| REDUALVAE | | 9.57 | **11.68** |
| ReHistoGAN | Logos[6] | 27.00 | 35.10 |
| REDUALVAE | | **9.31** | **11.76** |
| ReHistoGAN | Landscapes | **9.28** | 29.46 |
| cVAE | | 15.53 | 15.33 |
| REDUALVAE | | 9.64 | **10.35** |
| ReHistoGAN | Art | **10.17** | 49.89 |
| cVAE | | 33.12 | 31.21 |
| REDUALVAE | | 14.36 | **14.69** |

Table 4. Colour Transfer and Grayscale Colourisation Comparison on 128 × 128 images: FID is used for evaluation. Lower is better.

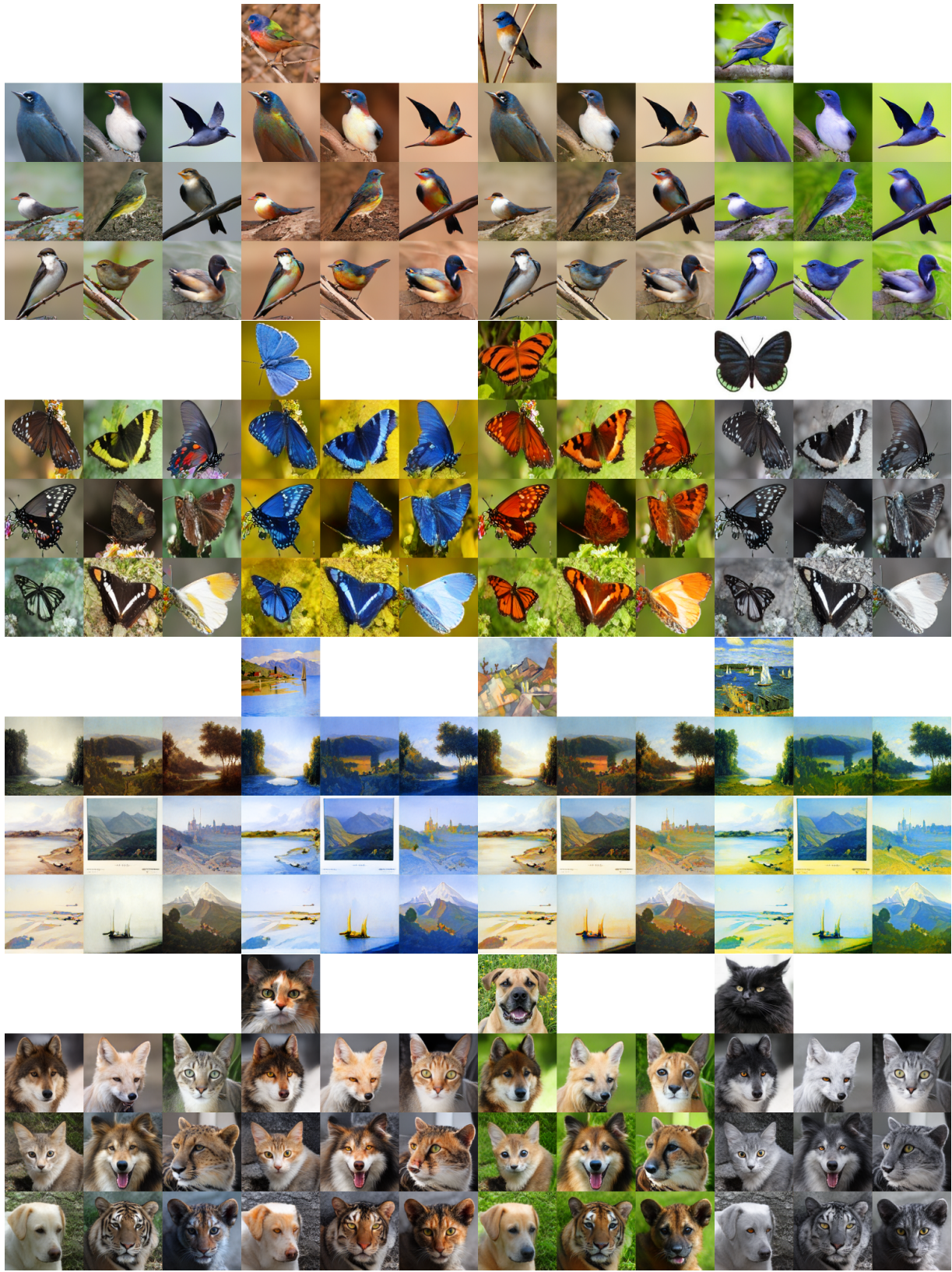

Fig. 6. Example image generation results of DualVAE on CUB_Birds, Butterfly, Art landscapes, Metfaces and AFHQ datasets. Each 3 × 3 block in the first columns are generated images under unconditional settings (i.e. both geometry and colour latents are sampled from the prior). However, the remaining four 3 × 3 blocks corresponds to conditional generation with colours of each of the 3 × 3 block derived from the exemplar directly above them. In all five blocks, the same nine geometric tokens are used.

## 6 ACKNOWLEDGEMENTS

The work has been supported by the Cyber Security Research Centre Limited whose activities are partially funded by the Australian Government's Cooperative Research Centres Program. This project was supported by resources and expertise provided by CSIRO IMT Scientific Computing.

## 7 APPENDIX

### 7.1 Notation and Definitions

| Notation | Definition |
|---|---|
| $X$ | Image instance |
| $F_g$ | Geometry features corresponding to $X$ |
| $F_c$ | Colour features corresponding to $X$ |
| $z_c$ | Latent variable of $F_c$ |
| $z_g$ | Latent variable of $F_g$ |
| $D_X$ | Decoder of image $X$ |
| $D_g$ | Decoder of geometry features $F_g$ |
| $D_c$ | Decoder of colour features $F_c$ |
| $E_X$ | Encoder of image $X$ |
| $E_g$ | Encoder of geometry features $F_g$ |
| $E_c$ | Encoder of colour features $F_c$ |
| $\theta_X$ | Parameters of the decoder $D_X$ |
| $\theta_g$ | Parameters of the decoder $D_g$ |
| $\theta_c$ | Parameters of the decoder $D_c$ |
| $\phi_X$ | Parameters of the encoder $E_X$ |
| $\phi_g$ | Parameters of the encoder $E_g$ |
| $\phi_c$ | Parameters of the encoder $E_c$ |
| $\|\ldots\|_1$ | $L_1$ norm of ... |
| $\mathbb{E}_p[\ldots]$ | Expectation of expression ... over probability density $p$ |
| $\mathbb{KL}(P, Q)$ | Kullback–Leibler divergence between probability distribution $P$ and $Q$ |
| $q_\phi$ | Approximate posterior whose model parameters is $\phi$ |
| $p_\theta$ | Probability density whose model parameters is $\theta$ |

### 7.2 Derivation details of DualVAE ELBO

Theorem 1. *If random variables $z_c, z_g, F_c, F_g$ and $X$ are such that their joint probability distribution satisfies*

$$p(X, F_g, F_c, z_c, z_g) = p(X|F_g, F_c) \cdot p(F_c|z_c) \cdot p(F_g|z_g) \cdot p(z_g) \cdot p(z_c)$$

*(or equivalently expressed by a probabilistic graphical model as in Figure 7) Then*

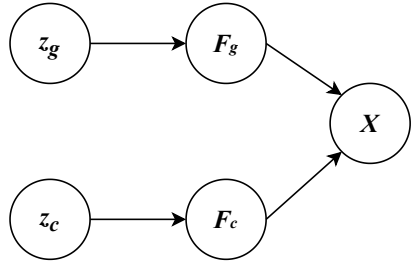

Fig. 7. Probabilistic Graphical Model of DualVAE

$$\begin{aligned}
\mathcal{L}_{\text{Explicit ELBO}} &\stackrel{\text{def}}{=} \mathbb{E}_{q_{\phi_c}(z_c|F_c) \cdot q_{\phi_g}(z_g|F_g)} \left[\log(p_{\theta_X}(X|F_g, F_c))\right] \\
&+ \mathbb{E}_{q_{\phi_g}(z_g|F_g)} \left[\log p_{\theta_g}(F_g|z_g)\right] - \mathbb{KL}\left[q_{\phi_g}(z_g|F_g) || p(z_g)\right] \\
&+ \mathbb{E}_{q_{\phi_c}(z_c|F_c)} \left[\log p_{\theta_i}(F_c|z_c)\right] - \mathbb{KL}\left[q_{\phi_c}(z_c|F_c) || p(z_c)\right]
\end{aligned} \quad (5)$$

*is a lower-bound of $\log p_\theta(X, F_g, F_c)$*

Proof. We lower bound $\log p_\theta(X, F_g, F_c)$ in the standard way using Jensen's inequality, but with separate approximate posterior distribution $q_{\phi_g}$ and $q_{\phi_c}$ for $z_g$ and $z_c$ respectively. Then $p_\theta(X, F_g, F_c)$ can be re-written based on the conditional independence assumption of the Theorem and marginalising over $z_g$ and $z_c$ results in

$$\log p_\theta(X, F_g, F_c) = \log \int p_{\theta_X}(X|F_g, F_c) \cdot p_{\theta_g, \theta_c}(F_g, F_c|z_g, z_c) \cdot p(z_g) \cdot p(z_c) \, dz_g dz_c$$

$$= \log \int \left\{ p_{\theta_g,\theta_c}(X|F_g, F_c) \cdot p_{\theta_g,\theta_c}(F_g, F_c|z_g, z_c) \cdot \frac{p(z_g) \cdot p(z_c)}{q_{\phi_g}(z_g|F_g) \cdot q_{\phi_c}(z_c|F_g)} \right\} \\ \times q_{\phi_g}(z_g|F_g) \cdot q_{\phi_c}(z_c|F_c) \, dz_g dz_c \quad (6)$$

Applying Jensen's inequality to Equation (6) and simplifying, we arrive at the ELBO of DualVAE given in Equation (??).

$$\log p_\theta(X, F_g, F_c) \geq \int \log \left\{ p_{\theta_g,\theta_c}(X|F_g, F_c) \cdot p_{\theta_g,\theta_c}(F_g, F_c|z_g, z_c) \cdot \frac{p(z_g) \cdot p(z_c)}{q_{\phi_g}(z_g|F_g) \cdot q_{\phi_c}(z_c|F_c)} \right\} \\ \times q_{\phi_g}(z_g|F_g) \cdot q_{\phi_c}(z_c|F_c) \, dz_g dz_c$$

$$= \int q_{\phi_g}(z_g|F_g) \cdot q_{\phi_c}(z_c|F_c) \cdot \log p_{\theta_g,\theta_c}(X|F_g, F_c) \quad dz_g, dz_c$$

$$+ \int q_{\phi_g}(z_g|F_g) \cdot q_{\phi_c}(z_c|F_c) \cdot \log p_{\theta_g,\theta_c}(F_g, F_c|z_g, z_c) \quad dz_g dz_c$$

$$- \int q_{\phi_g}(z_g|F_g) \cdot q_{\phi_c}(z_c|F_c) \cdot \log \left\{ \frac{q_{\phi_g}(z_g|F_g) \cdot q_{\phi_c}(z_c|F_c)}{p(z_g) \cdot p(z_c)} \right\} \quad dz_g dz_c \quad (7)$$

$$= \int q_{\phi_g}(z_g|F_g) \cdot q_{\phi_c}(z_c|F_c) \cdot \log p_{\theta_g,\theta_c}(X|F_g, F_c) \quad dz_g dz_c$$

$$+ \sum_{i=1}^{2} \int q_{\phi_g}(z_g|F_g) \cdot q_{\phi_c}(z_c|F_c) \cdot \log p_{\theta_i}(F_i|z_i) \quad dz_g dz_c$$

$$- \sum_{i=1}^{2} \int q_{\phi_g}(z_g|F_g) \cdot q_{\phi_c}(z_c|F_c) \cdot \log \left\{ \frac{q_{\phi_i}(z_i|F_i)}{p(z_i)} \right\} \quad dz_g dz_c \quad (8)$$

$$= \int q_{\phi_g}(z_g|F_g) \cdot q_{\phi_c}(z_c|F_c) \cdot \log p_{\theta_g,\theta_c}(X|F_g, F_c) \, dz_g dz_c$$

$$+ \int q_{\phi_g}(z_g|F_g) \cdot \log p_{\theta_g}(F_g|z_g) dz_g - \int q_{\phi_g}(z_g|F_g) \cdot \log \left( \frac{q_{\phi_g}(z_g|F_g)}{p(z_g)} \right) dz_g$$

$$+ \int q_{\phi_c}(z_c|F_c) \cdot \log p_{\theta_c}(F_c|z_c) dz_c - \int q_{\phi_c}(z_c|F_c) \cdot \log \left( \frac{q_{\phi_c}(z_c|F_c)}{p(z_c)} \right) dz_c \quad (9)$$

$$= \mathbb{E}_{q_{\phi_c}(z_c|F_c) \cdot q_{\phi_g}(z_g|F_g)} \left[ \log(p_{\theta_X}(X|F_g, F_c)) \right]$$

$$+ \mathbb{E}_{q_{\phi_g}(z_g|F_g)} \left[ \log p_{\theta_g}(F_g|z_g) \right] - \mathbb{KL} \left[ q_{\phi_g}(z_g|F_g) || p(z_g) \right]$$

$$+ \mathbb{E}_{q_{\phi_c}(z_c|F_c)} \left[ \log p_{\theta_i}(F_c|z_c) \right] - \mathbb{KL} \left[ q_{\phi_c}(z_c|F_c) || p(z_c) \right] \quad (10)$$

$\square$

### 7.3 Laplace distribution of image pixels

LEMMA 1. *If the un-normalised conditional probability density of $F_g$, $F_c$ and $F_X$ are*

$$p_{\theta_g}(F_g|z_g) \propto \exp\left(-\left\|F_g - D_g(z_g)\right\|_1\right) \quad (11)$$

$$p_{\theta_c}(F_c|z_c) \propto \exp\left(-\left\|F_c - D_c(z_c)\right\|_1\right) \quad (12)$$

$$p_{\theta_X}(X|F_g, F_c) \propto \exp\left(-\left\|X - D_c(F_g, F_c)\right\|_1\right) \quad (13)$$

*Then $\mathcal{L}_{\text{Explicit ELBO}}$ can be re-written as*

$$\mathcal{L}_{\text{Explicit ELBO}} = -\mathbb{E}_{q_{\phi_c}(z_c|F_c) \cdot q_{\phi_g}(z_g|F_g)} \left[ \left\| X - D_X(F_g, F_c) \right\|_1 \right] \quad (14)$$

$$- \mathbb{E}_{q_{\phi_g}(z_g|F_g)} \left[ \left\| F_g - D_g(z_g) \right\|_1 \right] - \mathbb{KL} \left[ q_{\phi_g}(z_g|F_g) || p(z_g) \right] \quad (15)$$

$$- \mathbb{E}_{q_{\phi_c}(z_c|F_c)} \left[ \left\| F_c - D_c(z_c) \right\|_1 \right] - \mathbb{KL} \left[ q_{\phi_c}(z_c|F_c) || p(z_c) \right] + \text{constant} \quad (16)$$

PROOF. We introduce a normalisation constant of $K$ for the conditional density of $F_g|z_g$ then we may re-write the expectation in-terms of $L_1$ norm resulting in

$$\mathbb{E}_{q_{\phi_g}(z_g|F_g)}\left[\log p_{\theta_g}(F_g|z_g)\right] = \mathbb{E}_{q_{\phi_g}(z_g|F_g)}\left[\log\left(\frac{1}{K} \cdot \exp\{-\|F_g - D_g(z_g)\|_1\}\right)\right] \quad (17)$$

$$= \mathbb{E}_{q_{\phi_g}(z_g|F_g)}\left[\log\left(\exp\{-\|F_g - D_g(z_g)\|_1\}\right)\right] + \text{constant}$$

$$= \mathbb{E}_{q_{\phi_g}(z_g|F_g)}\left[-\|F_g - D_g(z_g)\|_1\right] + \text{constant}$$

$$= -\mathbb{E}_{q_{\phi_g}(z_g|F_g)}\left[\|F_g - D_g(z_g)\|_1\right] + \text{constant} \quad (18)$$

A similar derivation can be made for the random variables $F_c|z_c$ and $X|F_g, F_c$. Substituting these re-written expectation into Equation (5) results in

$$\mathcal{L}_{\text{Explicit ELBO}} = -\mathbb{E}_{q_{\phi_c}(z_c|F_c) \cdot q_{\phi_g}(z_g|F_g)}\left[\|X - D_X(F_g, F_c)\|_1\right]$$

$$- \mathbb{E}_{q_{\phi_g}(z_g|F_g)}\left[\|F_g - D_g(z_g)\|_1\right] - \mathbb{KL}\left[q_{\phi_g}(z_g|F_g)\|p(z_g)\right]$$

$$- \mathbb{E}_{q_{\phi_c}(z_c|F_c)}\left[\|F_c - D_c(z_c)\|_1\right] - \mathbb{KL}\left[q_{\phi_c}(z_c|F_c)\|p(z_c)\right]$$

$$+ \text{constant}$$

□

## 7.4 Implicit ELBO

THEOREM 2. *If the decoder $D_X$ satisfies restrictive conditions such as in Lemma 2 Then*

$$\mathcal{L}_{\text{Explicit ELBO}} \geq \mathcal{L}_{\text{Implicit ELBO}} + \text{constant}$$

*where*

$$\mathcal{L}_{\text{Implicit ELBO}} \stackrel{\text{def}}{=} -2 \cdot \mathbb{E}_{q_{\phi_c}(z_c|F_c) \cdot q_{\phi_g}(z_g|F_g)} \left[ \left\| X - D_X(F_g, F_c) \right\|_1 \right]$$
$$- \mathbb{E}_{q_{\phi_c}(z_c|F_c) \cdot q_{\phi_g}(z_g|F_g)} \left[ \left\| X - D_X\left(D_g(z_g), D_c(z_c)\right) \right\|_1 \right]$$
$$- \mathbb{KL}\left[ q_{\phi_g}(z_g|F_g) || p(z_g) \right] - \mathbb{KL}\left[ q_{\phi_c}(z_c|F_c) || p(z_c) \right], \qquad (19)$$

PROOF. Consider the Explicit ELBO of Equation (16)

$$\mathcal{L}_{\text{Explicit ELBO}} = -\mathbb{E}_{q_{\phi_c}(z_c|F_c) \cdot q_{\phi_g}(z_g|F_g)} \left[ \left\| X - D_X(F_g, F_c) \right\|_1 \right]$$
$$- \mathbb{E}_{q_{\phi_g}(z_g|F_g)} \left[ \left\| F_g - D_g(z_g) \right\|_1 \right] - \mathbb{KL}\left[ q_{\phi_g}(z_g|F_g) || p(z_g) \right]$$
$$- \mathbb{E}_{q_{\phi_c}(z_c|F_c)} \left[ \left\| F_c - D_c(z_c) \right\|_1 \right] - \mathbb{KL}\left[ q_{\phi_c}(z_c|F_c) || p(z_c) \right] + \text{constant}$$

Then noting that

$$\mathbb{E}_{q_{\phi_g}(z_g|F_g)} \left[ \left\| F_g - D_g(z_g) \right\|_1 \right] + \mathbb{E}_{q_{\phi_c}(z_c|F_c)} \left[ \left\| F_c - D_c(z_c) \right\|_1 \right]$$
$$= \mathbb{E}_{q_{\phi_c}(z_c|F_c) \cdot q_{\phi_g}(z_g|F_g)} \left[ \left\| F_c - D_c(z_c) \right\|_1 + \left\| F_g - D_g(z_g) \right\|_1 \right], \qquad (20)$$

we re-write $\mathcal{L}_{\text{Explicit ELBO}}$ as

$$\mathcal{L}_{\text{Explicit ELBO}} = -\mathbb{E}_{q_{\phi_c}(z_c|F_c) \cdot q_{\phi_g}(z_g|F_g)} \left[ \left\| X - D_X(F_g, F_c) \right\|_1 \right]$$
$$- \mathbb{E}_{q_{\phi_c}(z_c|F_c) \cdot q_{\phi_g}(z_g|F_g)} \left[ \left\| F_c - D_c(z_c) \right\|_1 + \left\| F_g - D_g(z_g) \right\|_1 \right]$$
$$- \mathbb{KL}\left[ q_{\phi_g}(z_g|F_g) || p(z_g) \right] - \mathbb{KL}\left[ q_{\phi_c}(z_c|F_c) || p(z_c) \right] + \text{constant} \qquad (21)$$

Now under suitable conditions such as in Lemma 2

$$\left\| F_c - D_c(z_c) \right\|_1 + \left\| F_g - D_g(z_g) \right\|_1 \leq \left\| X - D_X\left(D_g(z_g), D_c(z_c)\right) \right\|_1 + \left\| X - D_X(F_g, F_c) \right\|_1$$

Therefore the second term of Equation (21) can be lower-bounded:

$$-\mathbb{E}_{q_{\phi_c}(z_c|F_c) \cdot q_{\phi_g}(z_g|F_g)} \left[ \left\| F_c - D_c(z_c) \right\|_1 + \left\| F_g - D_g(z_g) \right\|_1 \right] \geq$$
$$-\mathbb{E}_{q_{\phi_c}(z_c|F_c) \cdot q_{\phi_g}(z_g|F_g)} \left[ \left\| X - D_X\left(D_g(z_g), D_c(z_c)\right) \right\|_1 + \left\| X - D_X(F_g, F_c) \right\|_1 \right] \qquad (22)$$

Now substituting Equation (22) with Equation (21) results in a lower-bound of $\mathcal{L}_{\text{Explicit ELBO}}$ given by

$$\mathcal{L}_{\text{Explicit ELBO}} \geq -\mathbb{E}_{q_{\phi_c}(z_c|F_c) \cdot q_{\phi_g}(z_g|F_g)} \left[ \left\| X - D_X(F_g, F_c) \right\|_1 \right]$$
$$- \mathbb{E}_{q_{\phi_c}(z_c|F_c) \cdot q_{\phi_g}(z_g|F_g)} \left[ \left\| X - D_X\left(D_g(z_g), D_c(z_c)\right) \right\|_1 + \left\| X - D_X(F_g, F_c) \right\|_1 \right]$$
$$- \mathbb{KL}\left[ q_{\phi_g}(z_g|F_g) || p(z_g) \right] - \mathbb{KL}\left[ q_{\phi_c}(z_c|F_c) || p(z_c) \right] + \text{constant}.$$

Further simplifying results in

$$\mathcal{L}_{\text{Explicit ELBO}} \geq -2 \cdot \mathbb{E}_{q_{\phi_c}(z_c|F_c) \cdot q_{\phi_g}(z_g|F_g)} \left[ \left\| X - D_X(F_g, F_c) \right\|_1 \right]$$
$$- \mathbb{E}_{q_{\phi_c}(z_c|F_c) \cdot q_{\phi_g}(z_g|F_g)} \left[ \left\| X - D_X\left(D_g(z_g), D_c(z_c)\right) \right\|_1 \right]$$
$$- \mathbb{KL}\left[ q_{\phi_g}(z_g|F_g) || p(z_g) \right] - \mathbb{KL}\left[ q_{\phi_c}(z_c|F_c) || p(z_c) \right] + \text{constant}.$$

In other words,

$$\mathcal{L}_{\text{Explicit ELBO}} \geq \mathcal{L}_{\text{Implicit ELBO}} + \text{constant}$$

□

LEMMA 2. *If the function* $D_X : \mathbb{R}^M \to \mathbb{R}^N$ *has a* $C > 0$ *such that for all* $a, b$

$$\|a - b\|_1 \leq C \cdot \|D_X(a) - D_X(b)\|_1 \tag{23}$$

*(Also known as **reverse-Lipschitz**) then*

$$\begin{aligned}\|F_g - D_g(z_g)\|_1 + \|F_c - D_c(z_c)\|_1 \\ \leq C \cdot \left(\|D_X(F_g, F_c) - X\|_1 + \|D_X(D_g(z_g), D_c(z_c)) - X\|_1\right)\end{aligned} \tag{24}$$

PROOF. We first note that

$$\|F_g - D_g(z_g)\|_1 + \|F_c - D_c(z_c)\|_1 = \|[F_g, F_c] - [D_g(z_g), D_c(z_c)]\|_1 \tag{25}$$

Then from the reverse-Lipschitz

$$\begin{aligned}\|F_g - D_g(z_g)\|_1 + \|F_c - D_c(z_c)\|_1 = \|[F_g, F_c] - [D_g(z_g), D_c(z_c)]\|_1 \\ \leq C \cdot \|D_X(F_g, F_c) - D_X(D_g(z_g), D_c(z_c))\|_1\end{aligned} \tag{26}$$

We further bound by applying the triangle inequality of distances to arrive at

$$\begin{aligned}\|D_X(F_g, F_c) - D_X(D_g(z_g), D_c(z_c))\|_1 \\ \leq \|D_X(F_g, F_c) - X\|_1 + \|D_X(D_g(z_g), D_c(z_c)) - X\|_1\end{aligned} \tag{27}$$

Then using Equation (26) and Equation (27), we arrive at the equation

$$\|F_g - D_g(z_g)\|_1 + \|F_c - D_c(z_c)\|_1 \tag{28}$$

$$\leq C \cdot \left(\|D_X(F_g, F_c) - X\|_1 + \|D_X(D_g(z_g), D_c(z_c)) - X\|_1\right) \tag{29}$$

Motivated by our theoretical result, we propose to treat the Lipschitz constant as a hyper-parameter of our algorithm, which we set to $C = 1$. □

## 7.5 Optimisation Steps

Our objective is to learn model parameters that maximises the joint likelihood of $X, F_g$ and $F_c$;

$$\text{Maximise}_\theta \quad p_\theta(X, F_c, F_g) \tag{30}$$

where $\theta$ includes the parameters $\theta_g, \theta_c, \phi_g, \phi_c, \theta_X$. However, $p_\theta(X)$ is intractable, so we maximise Implicit ELBO shown in Equation (19). The training objective of DUALVAE is to

$$\text{Maximise}_\theta \quad \mathcal{L}_{\text{Implicit ELBO}} + \text{constant} \tag{31}$$

We may remove the constant that does not depend on our optimisation parameters. Hence the maximisation problem is equivalent to

$$\text{Maximise}_\theta - \mathbb{E}_{q_{\phi_c}(z_c|F_c) \cdot q_{\phi_g}(z_g|F_g)} \left[\|X - D_X(D_g(z_g), D_c(z_c))\|_1\right] \tag{32}$$

$$- \mathbb{KL}\left[q_{\phi_g}(z_g|F_g)||p(z_g)\right] - \mathbb{KL}\left[q_{\phi_c}(z_c|F_c)||p(z_c)\right] \tag{33}$$

$$- 2 \cdot \mathbb{E}_{q_{\phi_c}(z_c|F_c) \cdot q_{\phi_g}(z_g|F_g)} \left[\|X - D_X(F_g, F_c)\|_1\right] \tag{34}$$

The inference loss of our model shown in Equation (33) has two terms that are optimised separately. The codeboook inference loss given by

$$\mathbb{KL}\left[q_{\phi_g}(z_g|F_g)||p(z_g)\right]$$

is optimised as in the VQ-VAE [Van Den Oord et al. 2017] model, using both stop-gradient and exponentially moving average. The Gaussian inference loss

$$\mathbb{KL}\left[q_{\phi_c}(z_c|F_c)||p(z_c)\right]$$

is optimised using the reparameterization trick.

## 7.6 Data

| Data | Size | URL |
| --- | --- | --- |
| Flowers | 116,00 | https://www.kaggle.com/datasets/bogdancretu/flower299 |
| Flowers | 95,400 | https://www.kaggle.com/datasets/msheriey/104-flowers-garden-of-eden |
| Art | 97,349 | https://www.kaggle.com/datasets/ipythonx/wikiart-gangogh-creating-art-gan |
| LLD Logos | 122,920 | https://data.vision.ee.ethz.ch/sagea/lld/ |
| FFHQ[Karras et al. 2020b] | 70,000 | https://github.com/NVlabs/ffhq-dataset |
| Anime Faces | 400,000 | Subset of https://www.kaggle.com/datasets/muoncollider/danbooru2020, cleaned and aligned |
| Butterfly | 13,600 | https://www.kaggle.com/competitions/lnu-deep-learn-1-image-classification/data, |
| AFHQ | 16,130 | https://github.com/clovaai/stargan-v2/ |
| MetFaces | 2,621 | https://github.com/NVlabs/metfaces-dataset |

Table 5. Data sets used. Train and test split are 95% and 5%.

## 7.7 Hyperparameters

Our Hyperparameters are tabulated in Table 6 and no hyperparameter optimisation was performed. In addition, our networks starting weights were set by Kaiming Initialisation.

| Parameter | Value |
| --- | --- |
| Batch size | 8 |
| Optimizer | Adam |
| Adam: $\beta_1$ | 0.9 |
| Adam: $\beta_2$ | 0.999 |
| Adam: $\epsilon$ | 1e-8 |
| Adam: Learning rate | 0.0005 |
| Reconstruction Loss | L1 + LPIPS |

Table 6. Hyper-parameter settings for the experiment. In this paper, we have kept all hyper-parameters constant and did not perform any hyper-parameter optimisation.

### 7.7.1 Parameters of the Autoregressive model trained on the quantised codebook.
We used transformer model to train on the token sequence. The parameters used is shown in 7.

| Parameter | Value |
| --- | --- |
| Channel | 256 |
| Kernel size | 5 |
| Number of blocks | 4 |
| Number of residual blocks | 4 |
| Residual channel | 256 |
| Dropout | 0.1 |

Table 7. Parameters of Transformer prior

## 8 SUPPLEMENTARY RESULTS

*8.0.1 ReDualVAE: Uncurated Colour Transfer Results.*

*8.0.2 ReDualVAE: Recolouring Images.* In this section, we show colouring results obtained from ReDualVAEin Figure 13. We see that our model recognizes segments in the grayscale images and performs semantically meaningful colourisation of them.

*8.0.3 ReDualVAE: Interpolation in the colour latent space.*

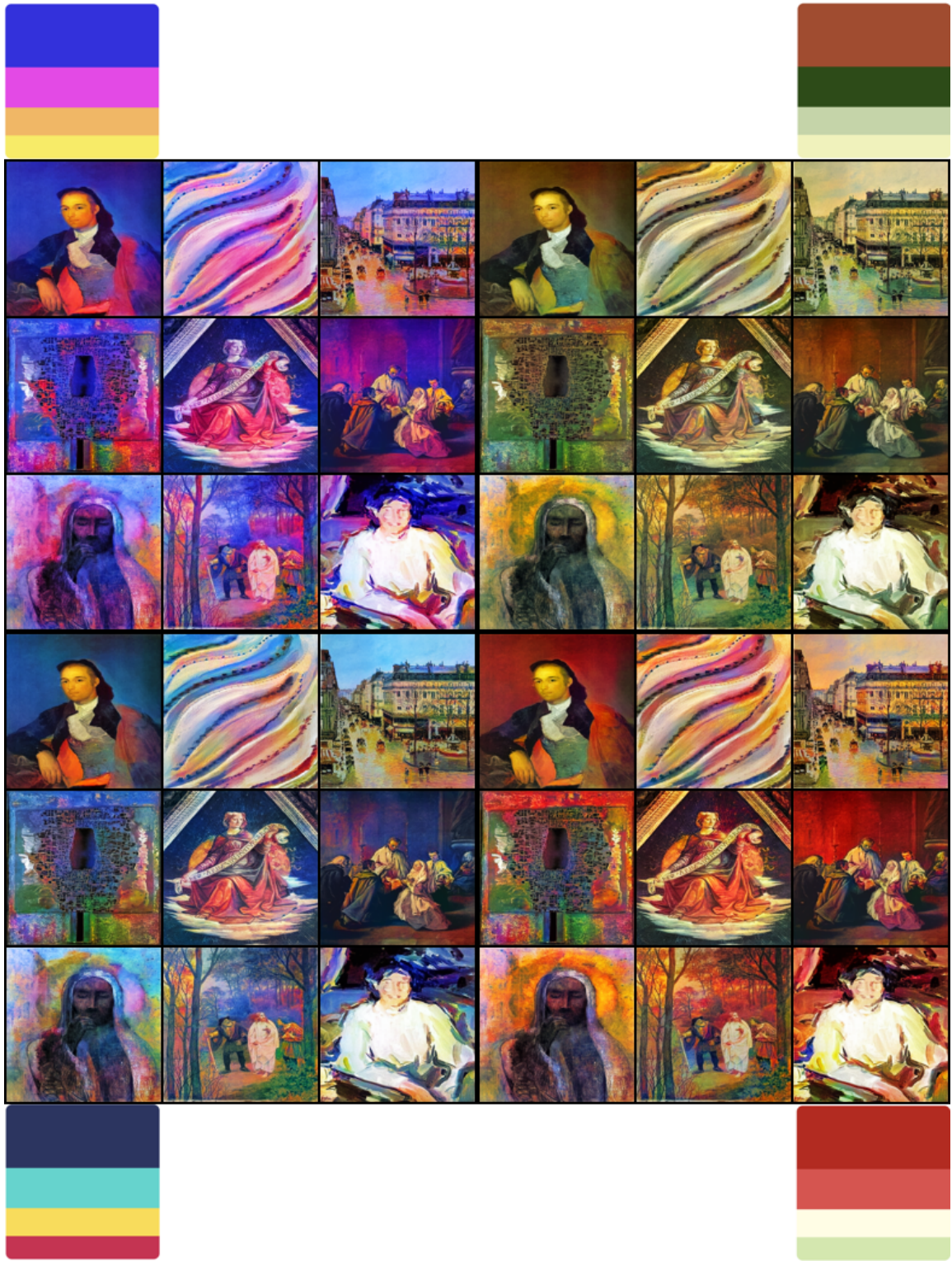

Fig. 8. Random selected colour transfer results of ReDualVAE on art dataset. Colours for each 3 × 3 quadrant are derived from the exemplar at the four corners. More example results are available in the supplementary.

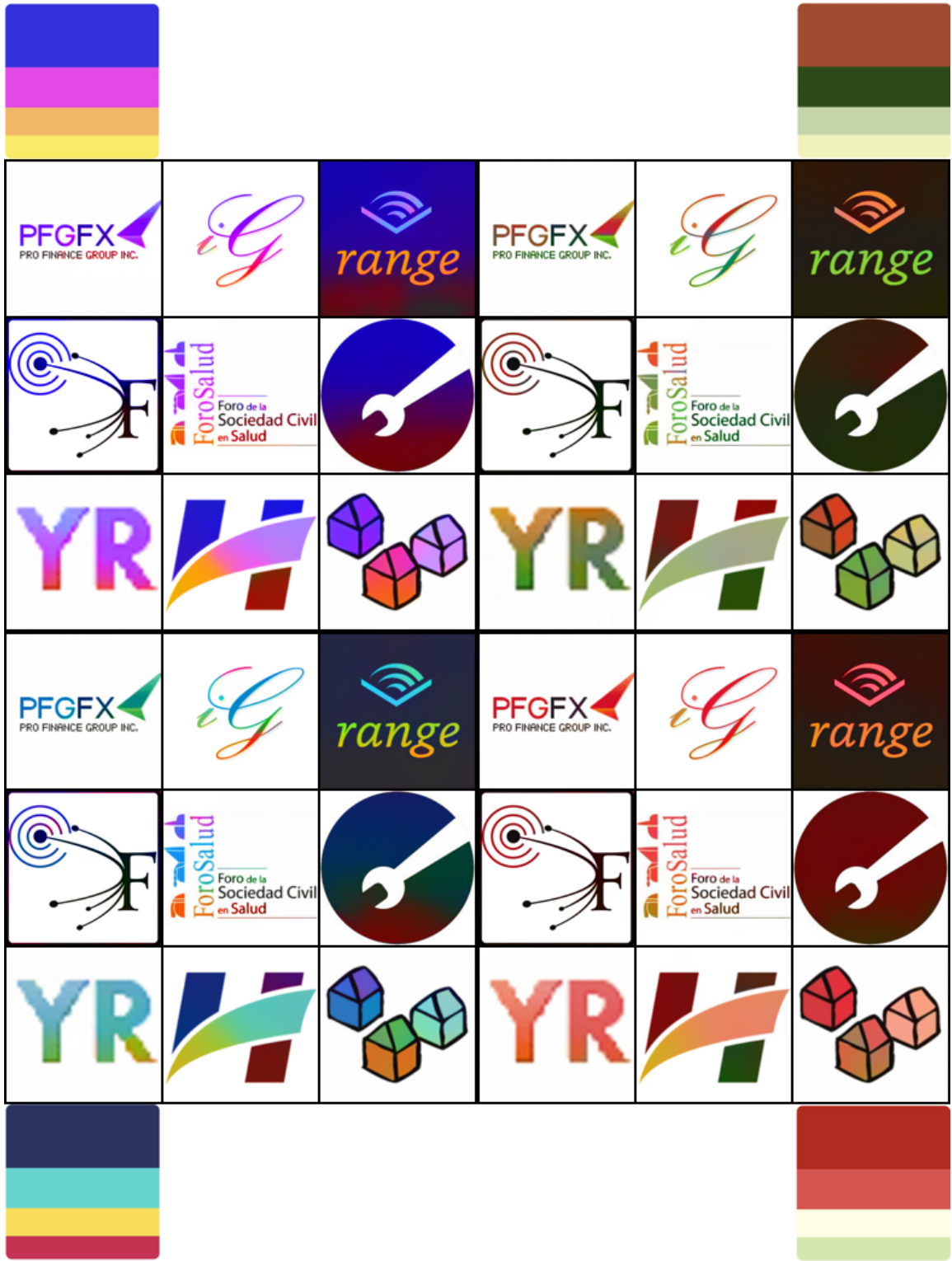

Fig. 9. Random selected colour transfer results of ReDualVAE on Logos dataset. Colours for each 3 × 3 quadrant are derived from the exemplar at the four corners. More example results are available in the supplementary.

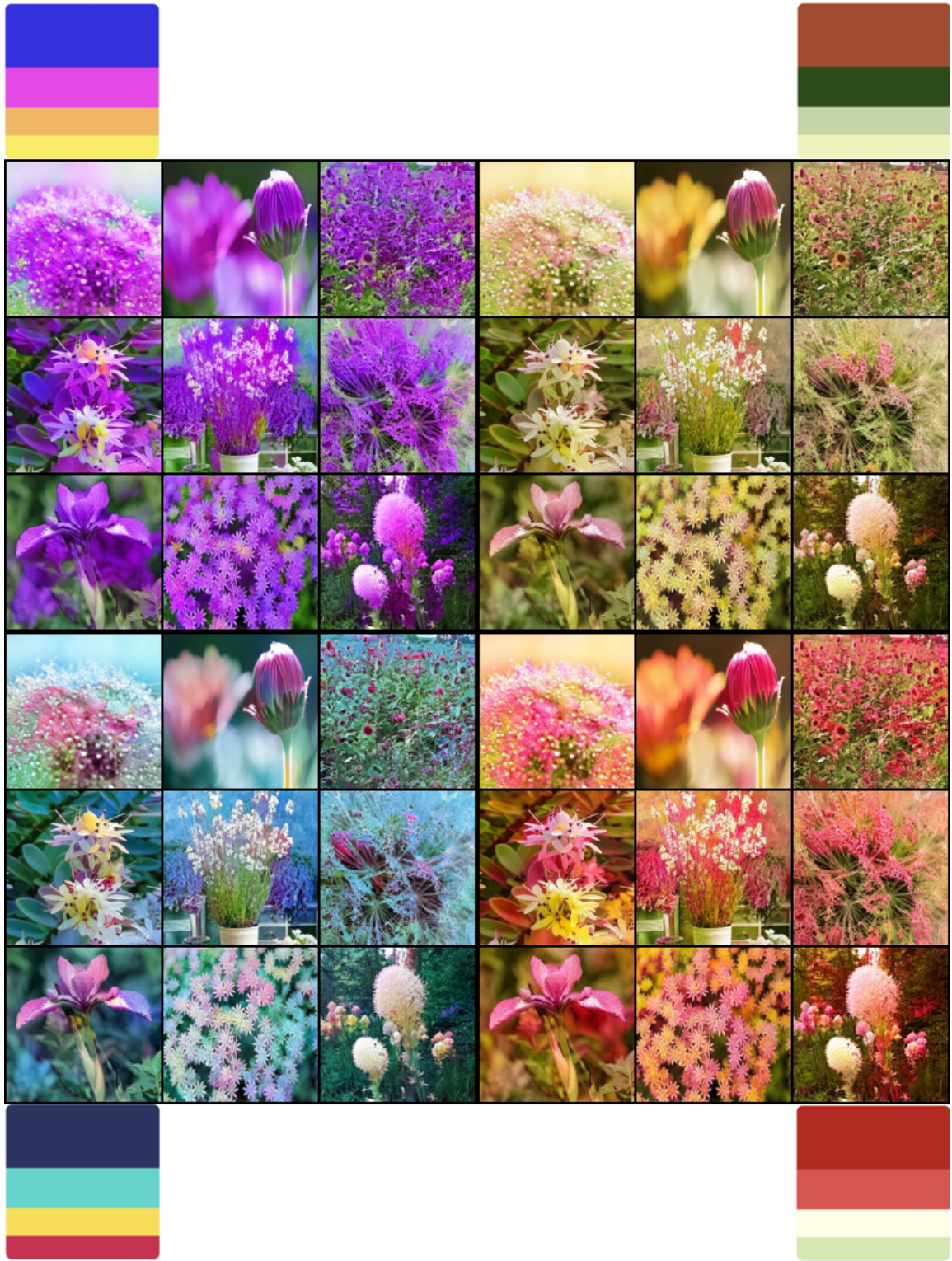

Fig. 10. Uncurated colour transfer results of ReDualVAE on Flowers dataset. Colours for each 3 × 3 quadrant are derived from the exemplar at the four corners. More example results are available in the supplementary.

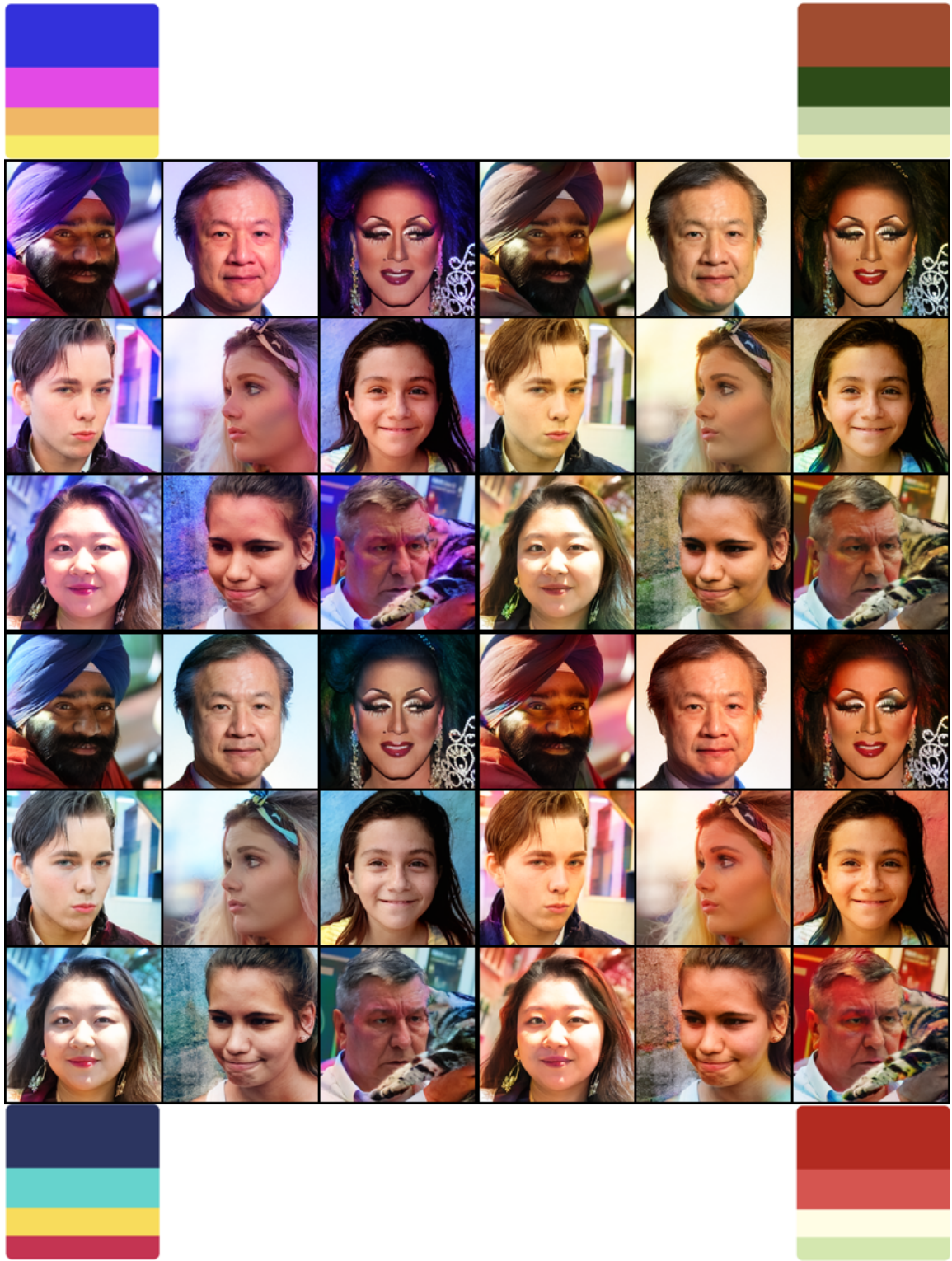

Fig. 11. Uncurated colour transfer results of ReDualVAE on FFHQ dataset. Colours for each 3 × 3 quadrant are derived from the exemplar at the four corners. More example results are available in the supplementary.

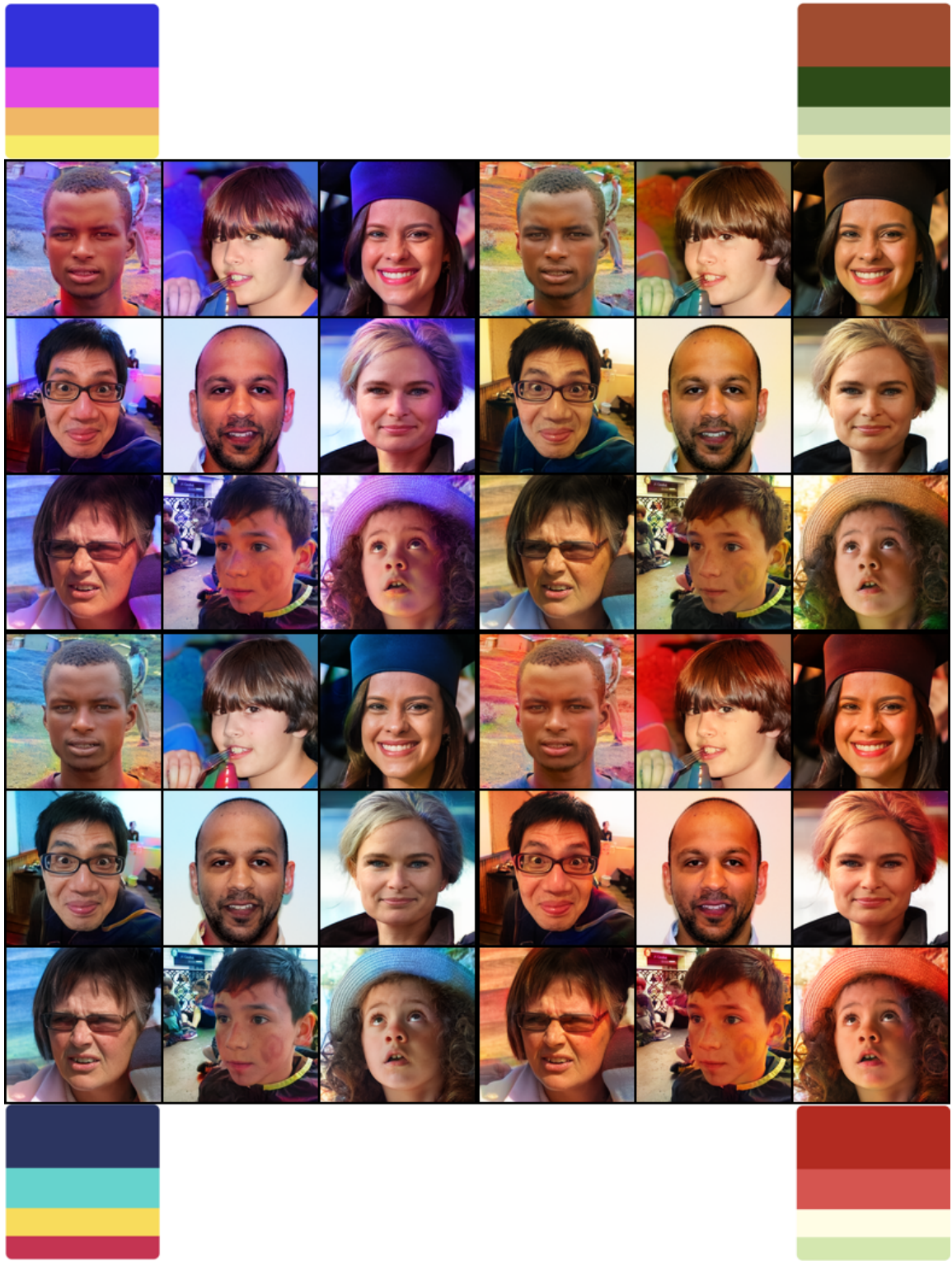

Fig. 12. Uncurated colour transfer results of ReDualVAE on FFHQ dataset. Colours for each 3 × 3 quadrant are derived from the exemplar at the four corners. More example results are available in the supplementary.

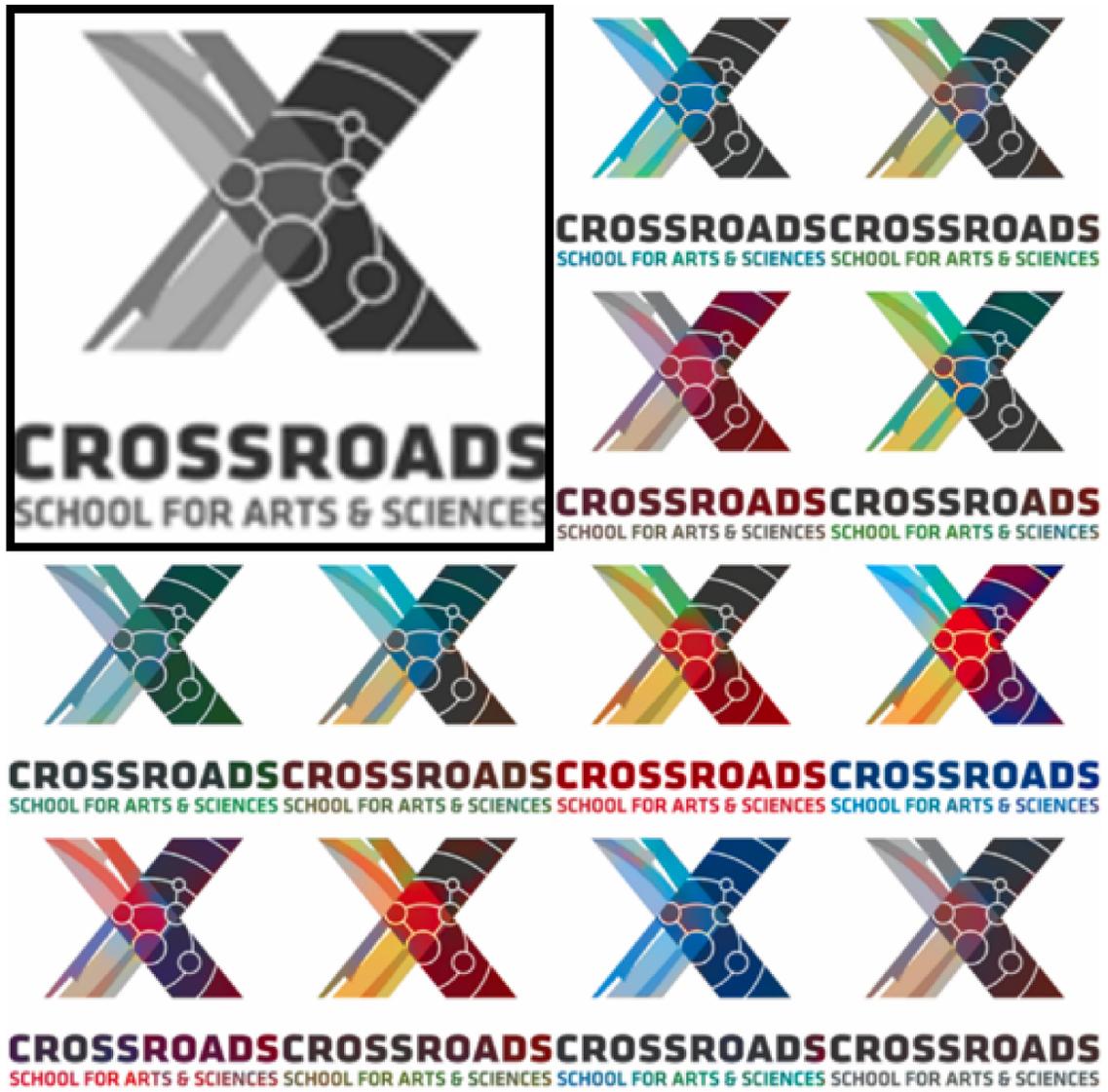

Fig. 13. Examples diverse re-colourisations from ReDualVAE on Logos. The gray image at the top-left is recoloured. Re-colorisation is performed by sampling from the Gaussian colour prior $z_c \sim \mathcal{N}(0, 1)$.

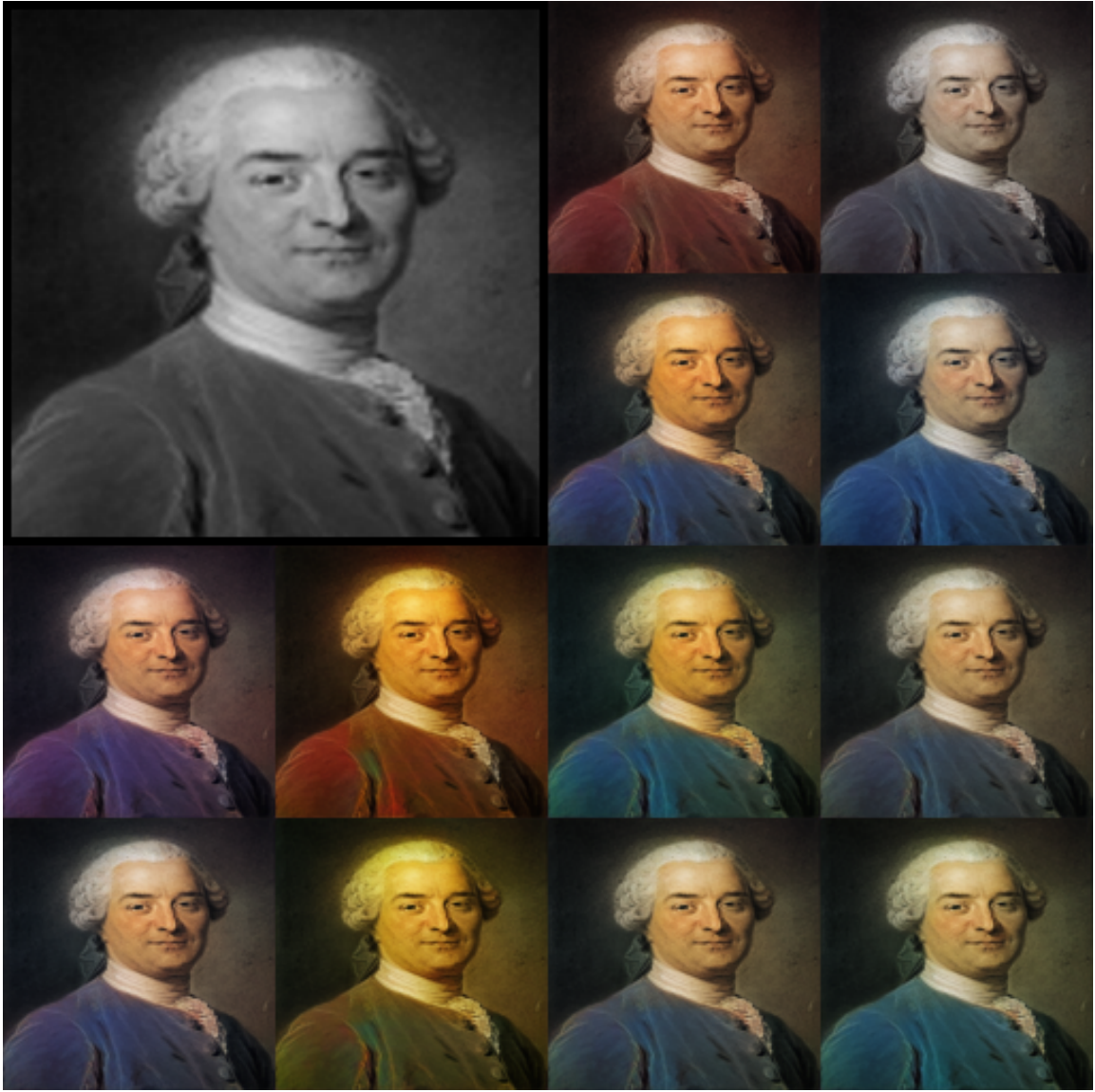

Fig. 14. Examples diverse re-colourisations from ReDualVAE on Art. The gray image at the top-left is recoloured. Re-colorisation is performed by sampling from the Gaussian colour prior $z_c \sim \mathcal{N}(0, 1)$.

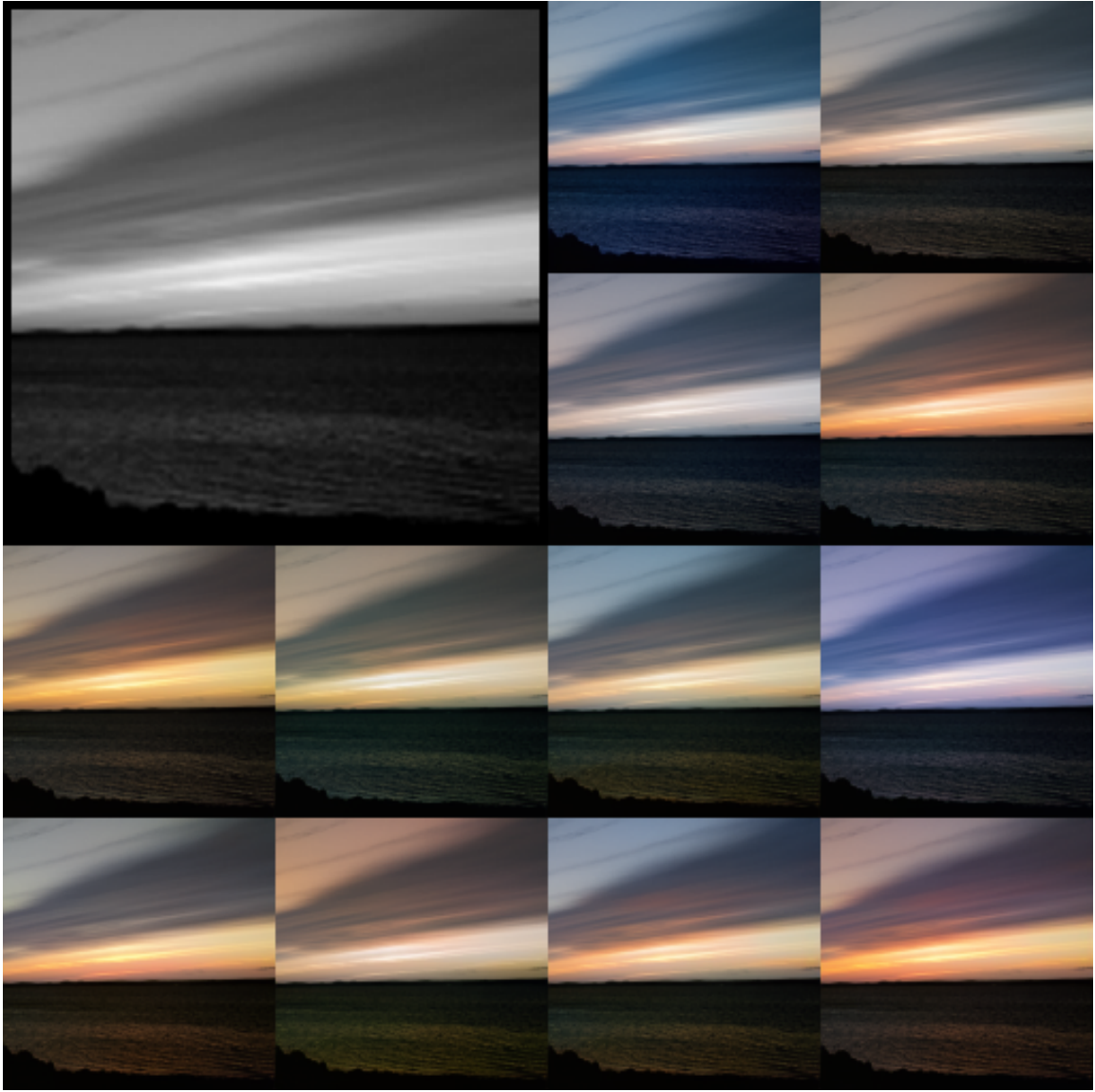

Fig. 15. Examples diverse re-colourisations from ReDualVAEon Landscapes. The gray image at the top-left is recoloured. Re-colorisation is performed by sampling from the Gaussian colour prior $z_c \sim \mathcal{N}(0, 1)$.

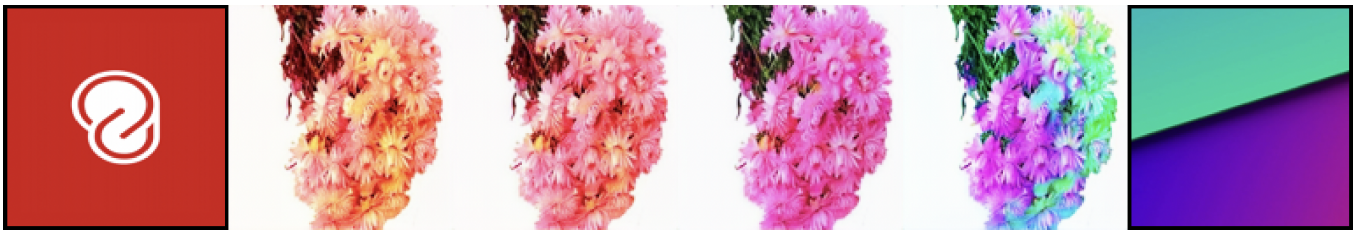

Fig. 16. Interpolation in the colour latent space leads to meaningful colourisation. The colours are derived from the exemplar images on the left and right. Interpolation is performed by increasing the weight of the right exemplars colour latent representation.